\documentclass[10pt,twocolumn,letterpaper]{article}

\usepackage{cvpr}
\usepackage{times}
\usepackage{epsfig}
\usepackage{graphicx}
\usepackage{amsmath}
\usepackage{amssymb}
\usepackage[ruled,vlined,linesnumbered]{algorithm2e}
\usepackage[noend]{algpseudocode}
\usepackage{enumerate}
\usepackage{comment}

\usepackage{xspace}

\newcommand{\Alg}{Alg.\xspace}

\newcommand{\htx}{\hat{x}}
\newcommand{\hty}{\hat{y}}
\newcommand{\calI}{\mathcal{I}}
\newcommand{\calU}{\mathcal{U}}

\usepackage[pagebackref=true,breaklinks=true,letterpaper=true,colorlinks,bookmarks=false]{hyperref}

\cvprfinalcopy 

\def\cvprPaperID{0366} 

\ifcvprfinal\fi
\begin{document}

\title{Gaussian Guided IoU: A Better Metric for Balanced Learning on Object Detection}

\author{Shengkai Wu\dag$^1$\quad Jinrong Yang\dag$^1$\quad Lijun Gou$^1$\quad Hangcheng Yu$^1$\quad Xiaoping Li*$^1$\\\\
	$^1$State Key Laboratory of Digital Manufacturing Equipment and Technology\\
	Huazhong University of Science and Technology, Wuhan, 430074, China.\\
	\dag Equal contribution\quad\quad\quad\quad\quad *Corresponding Author \\
{\tt\small \{ShengkaiWu, yangjinrong\}@hust.edu.cn}}

\maketitle
\begin{abstract}

For most of the anchor-based detectors, Intersection over Union(IoU) is widely utilized to assign targets for the anchors during training. However, IoU pays insufficient attention to the closeness of the anchor's center to the truth box's center. This results in two problems: (1) only one anchor is assigned to most of the slender objects which leads to insufficient supervision information for the slender objects during training and the performance on the slender objects is hurt; (2) IoU can not accurately represent the alignment degree between the receptive field of the feature at the anchor's center and the object. Thus during training, some features whose receptive field aligns better with objects are missing while some features whose receptive field aligns worse with objects are adopted. This hurts the localization accuracy of models. To solve these problems, we firstly design Gaussian Guided IoU(GGIoU) which focuses more attention on the closeness of the anchor's center to the truth box's center. Then we propose GGIoU-balanced learning method including GGIoU-guided assignment strategy and GGIoU-balanced localization loss. The method can assign multiple anchors for each slender object and bias the training process to the features well-aligned with objects. Extensive experiments on the popular benchmarks such as PASCAL VOC and MS COCO demonstrate GGIoU-balanced learning can solve the above problems and substantially improve the performance of the object detection model, especially in the localization accuracy.

\end{abstract}
\section{Introduction}
\label{introduction}


\begin{figure}[h]
\centering
\includegraphics[width=0.85\linewidth]{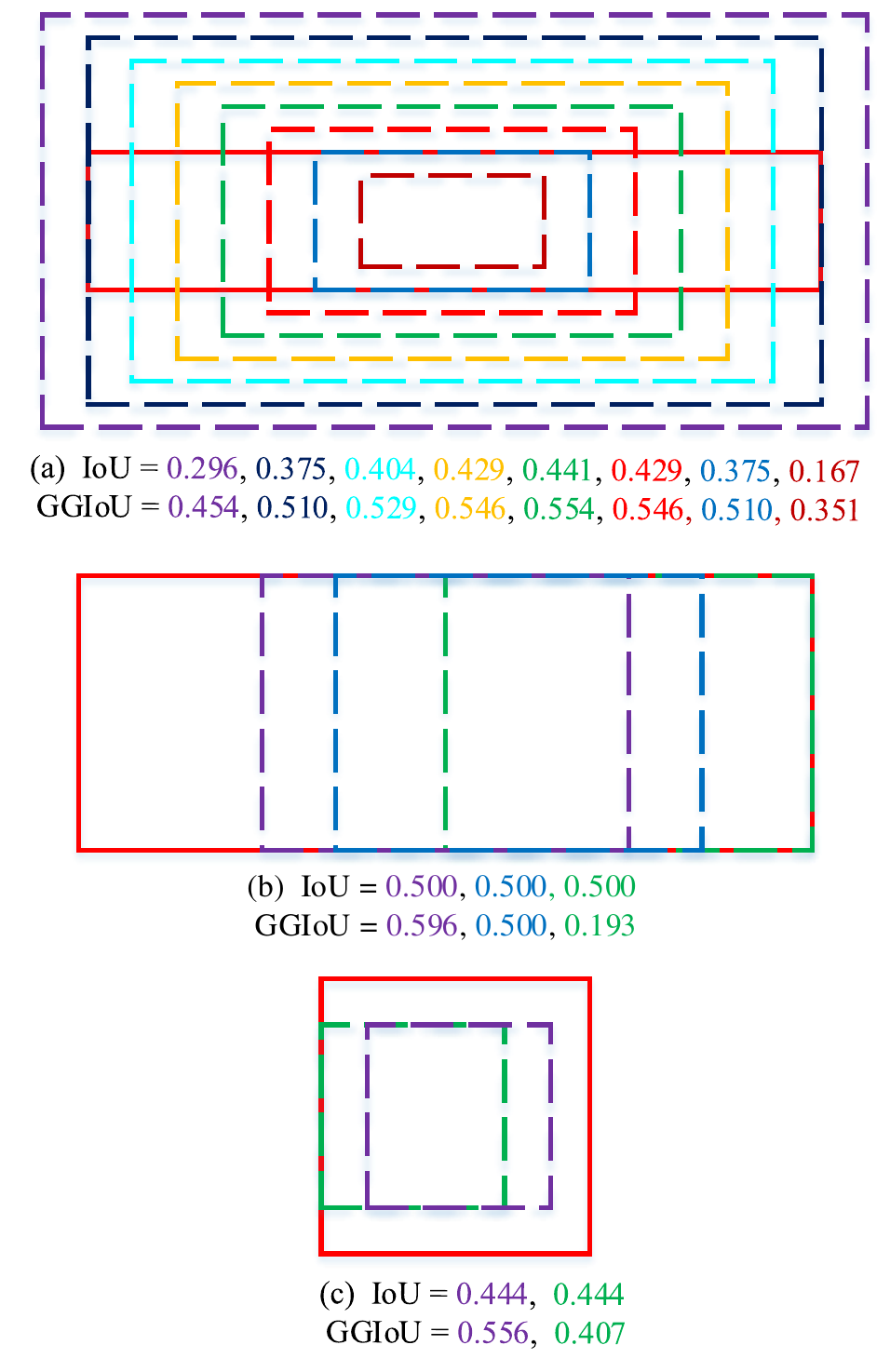}
\caption{Comparison between the impact of IoU and GGIoU. The boxes with solid lines represent truth boxes and the boxes with dashed lines represent anchors. (a) The IoU between the slender object and most of the anchors is small and only the nearest anchor can be assigned to the slender object when IoU is used to assign targets. But the GGIoU is large and multiple anchors can be assigned to the slender object when GGIoU is used to assign targets. (b) The receptive field of the feature at the purple anchor's center aligns better with the object than that of the features at the other anchors' centers. But IoU of these three anchors is the same and can not accurately represent the alignment degree of the features. The GGIoU of these anchors is different and can represent the alignment degree more accurately. (c) is similar to (b).}
\label{fig:GGIoU_analyse}
\end{figure}

With the advent of deep learning, object detection have developed quickly and many object detectors have been proposed. Most of object detectors can be classified into anchor-based detectors\cite{2017FasterRCNN, Liu2016SSD, fu2017dssd, Lin2017Focal, redmon2017yolo9000} and anchor-free detectors\cite{yu2016unitbox, redmon2016yolov1, tian2019fcos, kong2020foveabox}. Anchor-based detectors tile densely anchors with different scales and aspect ratios at each position on the feature map. For anchor-based detectors, Intersection over Union(IoU) is adopted as a metric to assign targets for anchors during training. IoU is the most popular metric for comparing the similarity between two arbitrary shapes due to its invariance of the scale of boxes. However, IoU pays insufficient attention to the closeness of the anchor's center to the truth box's center, resulting in two problems when utilizing IoU to assign targets.

\textbf{Firstly, only one anchor can be assigned to the truth box for most of the slender objects when IoU is used to assign targets.} RetinaNet\cite{Lin2017Focal} sets IoU threshold (0.4, 0.5) to assign positive examples and negative examples. When the IoU between an anchor and the nearest truth box is smaller than 0.4, this anchor is assigned to be negative example and when the IoU between an anchor and the nearest truth box is not smaller than 0.5, this anchor is assigned to the corresponding nearest truth box as positive example. Besides, RetinaNet assigns the truth box with the nearest anchors to ensure there exists at least one anchor for each truth box. Other anchor-based detectors such as Faster R-CNN\cite{2017FasterRCNN} and SSD\cite{Liu2016SSD} adopt a similar strategy to assign targets during training. With this assignment strategy, most of the slender objects can only be assigned with one anchor. Thus, there will be insufficient supervision information for training slender objects so that the trained models will perform badly for the slender objects. As Fig.\ref{fig:GGIoU_analyse}(a) shows, the red solid box is a slender object and the IoU between the nearby anchors and the truth box is below the positive threshold. Thus only the nearest green anchor can be assigned to the truth box when IoU is utilized to assign targets.

\textbf{Secondly, IoU can not accurately represent the alignment degree of the receptive field of the feature at the anchor's center.} Each anchor's center is always attached with a feature on the feature map. And the closeness of a feature to the truth box's center can represent the alignment degree between the receptive field of the feature and the truth box. The closer the feature is to the truth box's center, the better the receptive field of the feature aligns with the truth box. However, the IoU of anchors can not accurately represent the alignment degree of the receptive field of the corresponding feature. For example, as Fig.\ref{fig:GGIoU_analyse}(b), (c) show, the purple anchor's center are closer to the truth box's center than the green anchor's center. This reveals that the receptive field of the feature at the purple anchor's center aligns better than that of the feature at the green anchor's center. But the IoU of the purple anchor is the same as that of the green anchor. Thus, IoU fails to reveal the different alignment degree between the purple anchor and green anchor. During training, when IoU is used to assign targets, some features whose receptive field aligns better with the object are missing such as the feature at the purple anchor's center in Fig.\ref{fig:GGIoU_analyse}(c) while some features whose receptive field aligns worse with the object are adopted such as the feature at the green anchor's center in Fig.\ref{fig:GGIoU_analyse}(b), which can severely hurt the model's performance.

To solve these problems, we propose a better metric named Gaussian Guided IoU (GGIoU) to replace the standard IoU. We compute the Gaussian distance from each anchor's center to the truth box's center and then obtain GGIoU by multiplying the Gaussian distance with the standard IoU. Gaussian distance measures the closeness of the anchor's center to the truth box's center. The closer the anchor's center is to the truth box's center, the larger the Gaussian distance is. Thus, GGIoU pays more attention to the closeness of the anchor's center to the truth box's center than the standard IoU. We design GGIoU-balanced learning including GGIoU-guided assignment strategy and GGIoU-balnaced localization loss. This method makes the learning process more balanced and the above two problems can be solved. Extensive experiments on PASCAL VOC \cite{Everingham2010PascalVOC} and MS COCO \cite{lin2014MSCOCO} demonstrate that GGIoU-balanced learning can substantially improve the performance of object detectors, especially in the localization accuracy.

The main contribution of the paper is summarized as follows:
\begin{itemize}
  \setlength{\itemsep}{1pt}
  \setlength{\parskip}{0pt}
  \setlength{\parsep}{0pt}
 \item We design GGIoU by incorporating Gaussian distance into IoU as a better metric for comparing the similarity between two arbitrary shapes.
 \item We design GGIoU-balanced learning method including GGIoU-guided assignment strategy and GGIoU-balanced localization loss, which makes the detectors more powerful for accurate localization.
 \item We conduct extensive experiments on the popular benchmarks to demonstrate that GGIoU-balanced learning method can substantially improve the performance of the popular object detectors.
\end{itemize}

\section{Related Work}
\label{related work}
\textbf{IoU-based evaluation metric.} UnitBox\cite{yu2016unitbox} directly uses IoU loss to regresses the four bounds of a predicted box as a whole unit. GIoU\cite{2020GIoU} loss is proposed to tackle the issues of gradient vanishing for non-overlapping cases, but is still facing the problems of slow convergence and inaccurate regression. In comparison, DIoU and CIoU losses\cite{2020Distance-IoU} is proposed to perform faster convergence and better regression accuracy. GIoU, CIoU and DIoU is designed for the localization loss and can't be used for target assignment. Differently, GGIoU is mainly designed for the target assignment.

\textbf{Target assignment strategies in anchor-based detectors.} Anchor-based detectors commonly utlize IoU to assign targets. RetinaNet\cite{Lin2017Focal} assigns the anchors with IoU below the negative threshold as negative examples and the anchors with IoU not smaller than the positive threshold as positive examples. Faster RCNN\cite{2017FasterRCNN} and SSD\cite{Liu2016SSD} adopt the similar strategy. FreeAnchor\cite{zhang2019freeanchor} constructs a bag of anchors for each truth box based on the IoU and proposes detection customized loss for learning to match the best anchors. Similarly, MAL\cite{ke2020multipleanchorlearning} constructs bag anchors for each truth box and select the most representative anchor based on the classification and localization score. Besides, anchor depression is designed to perturb the features of selected anchors to decrease their confidence. ATSS\cite{zhang2020bridging} select K anchors from each level of FPN for each truth box and compute IoU for these anchors. Then the IoU threshold is adaptively computed based on the IoU distribution for each truth box. Differently, we design GGIoU as an alternative and propose GGIoU-guided assignment strategy.

\textbf{Feature alignment in object detection.} Feature alignment is important for the performance of object detection. Multi-stage detectors\cite{2017FasterRCNN,he2017maskrcnn,cai2018cascadercnn} adopt RoI pooling to extract the aligned feature for each proposal. Deformable convolution\cite{dai2017deformableConv,zhu2019deformablev2} predict the offset for each position in the convolution kernels and the receptive field of kernels can adaptively change based on the input feature. This makes feature align better with the object. Guided anchoring\cite{wang2019guidedanchoring} utilize the predicted box from the first stage to predict the offsets of deformable convolution and the deformable convolution can generate well-aligned feature for the second stage detector. AlignDet\cite{chen2019RoIConv} predict proposals at each position in the first stage and densely adopt RoIConv to extract the aligned region feature. Most of the models change the method of feature extraction to extract aligned feature. Differently, GGIoU-balanced learning make the training process focus more on the features close to the truth box center, which aligns better with objects.

\section{GGIoU-balanced Learning Method}

\subsection{Gaussian Guided Intersection over Union}
\label{section 3.1}

For anchor-based detectors, Intersection over Union (IoU) is widely used to assign targets for the anchors because of its appealing property of scale-invariant. IoU is computed by:

\begin{equation}\label{eq:iou}
IoU = \frac{|B^a\cap B^{gt}|}{|B^a\cup B^{gt}|},
\end{equation}
where $B^a = (x^a_1,y^a_1,x^a_2,y^a_2)$ is the coordinates of the top-left and bottom-right corners for the anchor box, and $B^{gt}=(x^{gt}_1,y^{gt}_1,x^{gt}_2,y^{gt}_2)$ is the coordinates of the top-left and bottom-right corners for the ground truth box. As analyzed above, there exist two problems when utilizing the standard IoU to assign targets for the anchors during training. 

\begin{algorithm}[!t]\label{algo:ggiou}
\caption{Gaussian Guided Intersection over Union (GGIoU)}
	\small{
	\SetKwInOut{Input}{input}\SetKwInOut{Output}{output}
		\Input{Parameter: $\beta$,\\
		anchor box $B^a$: $B^a = (x^a_1,y^a_1,x^a_2,y^a_2)$,\\
		ground truth box $B^{gt}$:\\
		$B^{gt}=(x^{gt}_1,y^{gt}_1,x^{gt}_2,y^{gt}_2)$.}
		\Output{{GGIoU}.}}
		For the anchor $B^a$, ensuring  $x^a_2>x^a_1$ and $y^a_2>y^a_1$:
		$\htx^a_1 = \min(x^a_1,x^a_2)$, $\quad \htx^a_2 = \max(x^a_1,x^a_2)$, $\quad\hty^a_1 = \min(y^a_1,y^a_2)$, $\quad\hty^a_2 = \max(y^a_1,y^a_2)$.\\
		Calculating area of $B^{gt}$: $S^{gt} = (x^{gt}_2 - x^{gt}_1)\times(y^{gt}_2 - y^{gt}_1)$.\\
		Calculating area of $B^a$: $\qquad \qquad S^a = (\htx^a_2 - \htx^a_1)\times(\hty^a_2 - \hty^a_1)$.\\
		Calculating intersection $\calI$ between $B^a$ and $B^{gt}$:
		$x^{\calI}_1 = \max(\htx^a_1,x^{gt}_1)$, $\quad x^{\calI}_2 = \min(\htx^a_2,x^{gt}_2)$, $\quad y^{\calI}_1 = \max(\hty^a_1,y^{gt}_1)$, $\quad y^{\calI}_2 = \min(\hty^a_2,y^{gt}_2)$,  $ w^{\calI} = \begin{cases} 
	    (x^{\calI}_2 - x^{\calI}_1) & \text{if}\quad x^{\calI}_2 > x^{\calI}_1\\
	    0 & \text{otherwise.}
        \end{cases}$, $\quad h^{\calI} = \begin{cases} 
        (y^{\calI}_2 - y^{\calI}_1) & \text{if}\quad y^{\calI}_2 > y^{\calI}_1\\
        0 & \text{otherwise.}
        \end{cases}$, $\qquad \qquad \calI = w^{\calI} \times h^{\calI}$.\\
		$\displaystyle IoU = \frac{\calI}{\calU}$, where  $\calU = S^a+S^{gt}-\calI$.\\
		Calculating the center coordinates of $B^a$: $\qquad \quad x_{c}^{a}=\frac{x_{1}^{a}+x_{2}^{a}}{2}$, $\quad y_{c}^{a}=\frac{y_{1}^{a}+y_{2}^{a}}{2}$.\\
		Calculating the center coordinates of $B^{gt}$: $x_{c}^{gt}=\frac{x_{1}^{gt}+x_{2}^{gt}}{2}$, $\quad y_{c}^{gt}=\frac{y_{1}^{gt}+y_{2}^{gt}}{2}$.\\
		Calculating Gaussian variance: $\qquad \qquad \qquad{{\sigma}_{1}}=\beta \times w^{\calI}$, $\quad{{\sigma}_{2}}=\beta \times h^{\calI}$.\\
		Calculating the Gaussian distance between the centers of truth box and anchor box $D_{c}$: ${{D}_{c}}={{e}^{-\frac{1}{2}\left( \frac{{{\left( x_{c}^{a}-x_{c}^{gt}\right)}^{2}}}{{{\sigma }_{1}}^{2}}+\frac{{{\left( y_{c}^{a}-y_{c}^{gt} \right)}^{2}}}{{{\sigma }_{2}}^{2}} \right)}}$.\\
		$\displaystyle GGIoU=Io{{U}^{\left( 1-\alpha  \right)}}{{D}_{c}}^{\alpha}$.
\end{algorithm}

To address these problems, we design a better metric named GGIoU to replace the standard IoU. We compute Gaussian distance $D_c$ from each anchor's center to the truth box's center and obtain Gaussian Guided IoU (GGIoU) by multiplying Gaussian distance with the standard IoU as Eq.\ref{eq:Gaussian distance} and Eq.\ref{eq:GGIoU} show.

\begin{equation}\label{eq:Gaussian distance}
{{D}_{c}}={{e}^{-\frac{1}{2}\left( \frac{{{\left( x_{c}^{a}-x_{c}^{gt}\right)}^{2}}}{{{\sigma }_{1}}^{2}}+\frac{{{\left( y_{c}^{a}-y_{c}^{gt} \right)}^{2}}}{{{\sigma }_{2}}^{2}} \right)}},
\end{equation}

\begin{equation}\label{eq:Gaussian variance}
{{\sigma}_{1}}=\beta \times w^{\calI},\quad{{\sigma}_{2}}=\beta \times h^{\calI},
\end{equation}

\begin{equation}\label{eq:GGIoU}
GGIoU=Io{{U}^{\left( 1-\alpha  \right)}}{{D}_{c}}^{\alpha },
\end{equation}

$(x_c^a, y_c^a)$ and $(x_c^{gt}, y_c^{gt})$ represent the coordinates of the anchor box's center and the truth box's center respectively. ${\sigma}_1$ and ${\sigma}_2$ represent the standard deviation in the $x$ and $y$ directions respectively and are computed by multiplying the parameter $\beta$ with the width $w^{\calI}$ and height $h^{\calI}$ of the overlap box respectively. Gaussian distance $D_c$ measures the closeness of the anchor's center to the truth box's center. The closer the anchor's center is to the truth box's center, the larger the Gaussian distance is. The range of $D_c$ is (0, 1]. Adopting the width $w^{\calI}$ and height $h^{\calI}$ of the overlap box to compute ${\sigma}_1$ and ${\sigma}_2$ is an important design. Assuming that there exist two anchors whose centers are the same but the size of overlap boxes are different, the anchor with a larger overlap box will get a larger standard deviation and thus get larger $D_c$, vice versa. Thus, Gaussian distance $D_c$ not only represents the closeness of the anchor's center to the truth box's center but also contains the size information of the overlap box between anchors and truth boxes. How about adopting the width and height of the truth box to compute the standard deviation ${\sigma}_1$ and ${\sigma}_2$? Obviously, $D_c$ of the anchor with a larger overlap box is the same as that of the anchor with a smaller overlap box under this condition and the size information of the overlap box is missing. Thus, adopting the overlap box to compute the standard deviation is more reasonable and this will be demonstrated in the subsequent experimental results. Compared with IoU, GGIoU focuses more attention on the closeness of the anchor box's center to the truth box's center. \Alg~\ref{algo:ggiou} describes the specific computation process of GGIoU. 

\subsection{GGIoU-guided Assignment Strategy}
\label{section 3.2}

To solve the above two problems, we design GGIoU-guided assignment strategy, which utilizes GGIoU to assign targets. Firstly, we compute the GGIoU between all the anchors and truth boxes. Then the anchors whose GGIoU with the nearest truth box is smaller than the negative threshold are assigned to be negative examples. The anchors whose GGIoU with the nearest truth box is not smaller than the positive threshold are assigned to be positive examples and are assigned to the nearest truth box. Finally, the truth boxes are assigned with the nearest anchors to ensure that all the truth boxes can be assigned with at least one anchor. GGIoU-guided assignment strategy can solve the above problem in the following ways. 

Firstly, as analyzed above, only one anchor can be assigned to the truth box for most of the slender objects when the standard IoU is used to assign targets. This problem can be solved by GGIoU-guided assignment strategy. GGIoU is influenced by both IoU and Gaussian distance $D_c$. For the anchors that are close to the center of the slender object, the Gaussian distance $D_c$ is large and can result in larger GGIoU even if IoU is small. As shown in Fig.\ref{fig:GGIoU_analyse}(a), the IoU of anchors (dotted boxes) with the slender object (red solid box) is quite small (0.296, 0.375, 0.404, 0.429, 0.441, 0.429, 0.375, 0.167) and only the nearest green anchor can be assigned to the slender object. However, the GGIoU of these anchors is larger (0.454, 0.510, 0.529, 0.546, 0.554, 0.546, 0.510, 0.351) because these anchors are at the center of the truth box. When GGIoU is utilized to assign targets, multiple anchors will be assigned to this slender object and more supervision information will be provided to train the model for these slender objects. Thus, the performance on the slender objects is improved.

Secondly, IoU can not accurately represent the alignment degree of the feature at the anchor's center with the truth box, which leads to that some features that align badly with the objects are adopted while some features that align well with the objects are missing during training. This problem can be alleviated by GGIoU-guided assignment strategy. As Fig.\ref{fig:GGIoU_analyse}(b) shows, the IoU of the purple anchor is the same as that of the green anchor and they all can be assigned to the truth box. But obviously, the receptive field of the feature at the purple anchor's center aligns better than that of the features at the green anchor's center. Thus, IoU can not accurately represent the alignment degree of the features. When IoU is utilized to assign targets, the feature at the green anchor's center which aligns worst with the object is adopted for training the object. But when GGIoU is used, the GGIoU of the green anchor is smaller than IoU and this green anchor is discarded. As Fig.\ref{fig:GGIoU_analyse}(c) shows, when IoU is used to assign targets, both the green and purple anchors are assigned to be negative examples and the feature at the purple anchor's center which aligns well with the object is missing. But when GGIoU is used, the central purple anchor is assigned to be a positive example and the corresponding feature is adopted for training this object. In summary, GGIoU-guided assignment strategy makes the training process bias to the features aligning well with objects (the purple anchor in Fig.\ref{fig:GGIoU_analyse}(c)) while discarding the features aligning badly with objects (the green anchor in Fig.\ref{fig:GGIoU_analyse}(b)). 

There still exists a problem. As Fig.\ref{fig:GGIoU_analyse}(b) shows, the feature at the blue anchor's center aligns worse than the feature at the purple anchor's center. No matter IoU or GGIoU is used, the blue anchor can be assigned to be a positive example and the feature at the purple anchor's center is treated the same as that at the purple anchor's center. Can we bias the model more to the features aligning better with the objects? Inspired by IoU-balanced localization loss\cite{iou-balanced}, we design GGIoU-balanced localization loss to realize this goal.

\subsection{GGIoU-balanced Localization Loss}
\label{section 3.3}

Under the framework of traditional training method, all target samples are equally treated by sharing the same weight in localization loss function. To guide the model pay more attention to the features aligning better with the objects, we proposed GGIoU-balanced localization loss as Eq.\ref{eq:GGIoU-balancedLoc} and Eq.\ref{eq:GGIoU-balancedLocPara} shows: 

\begin{equation}
 \label{eq:GGIoU-balancedLoc}
 {{L}_{loc}}=\sum\limits_{i\in Pos}^{N}{\sum\limits_{m\in cx,cy,w,h}{{{w}_{i}}(GGIo{{U}_{i}})*\text{smoot}{{\text{h}}_{L1}}(l_{i}^{m}-\hat{g}_{i}^{m})}}
\end{equation}

\begin{equation}
 \label{eq:GGIoU-balancedLocPara}
 {{w}_{i}}(GGIo{{U}_{i}})={{w}_{loc}}*GGIoU_{i}^{\lambda }
\end{equation}
where $(l_i^{cx}, l_i^{cy}, l_i^w, l_i^h)$ represents the encoded coordinates of prediction box and $(\hat{g}_i^{cx}, \hat{g}_i^{cy}, \hat{g}_i^w, \hat{g}_i^h)$ represents the encoded coordinates of truth boxes, which is the same as the decoding style of Faster R-CNN\cite{2017FasterRCNN}. ${w}_{loc}$ represents the weight of localization loss which is manually adjusted to ensure the loss in the first iteration the same as the baseline counterpart. 

There are two merits when using GGIoU-balanced localization function: (1) GGIoU-balanced localization loss can suppress the gradient of outliers during training; (2) GGIoU-balanced localization loss make the training process pay more attention to the features aligning better with objects.


\section{Experimental Results}
\label{results}

\begin{table}\footnotesize
\centering 
   \caption{\footnotesize The results of sampling positive and negative anchors based on GGIoU using the width and height of ground truths to export the standard deviation of Gaussian distances. $\alpha$ = 0 means the baseline model using IoU to sampling.}
 \begin{tabular}{cc|cccccc} 
 \hline
$\beta$ & $\alpha$ & mAP & $\text{AP}_{50}$ &$\text{AP}_{60}$ & $\text{AP}_{70}$ &$\text{AP}_{80}$ &$\text{AP}_{90}$\\
 \hline
  - & 0   & 0.308 & 0.496 & 0.450 & 0.376 & 0.261 & 0.086\\
1/6 & 0.4 & 0.309 & 0.487 & 0.448 & 0.377 & 0.270 & 0.094\\
1/6 & 0.3 & 0.313 & 0.492 & 0.451 & 0.381 & 0.271 & 0.095\\
1/6 & 0.2 & 0.309 & 0.490 & 0.448 & 0.377 & 0.268 & 0.089\\
1/5 & 0.4 & 0.306 & 0.485 & 0.444 & 0.374 & 0.264 & 0.090\\
1/5 & 0.3 & 0.313 & 0.496 & 0.455 & 0.384 & 0.271 & 0.094\\
1/5 & 0.2 & 0.312 & 0.493 & 0.452 & 0.383 & 0.268 & 0.093\\
 \hline
 \end{tabular}
  \label{table:4.1}
\end{table}

\begin{table}\footnotesize
\centering 
   \caption{\footnotesize The results of sampling positive and negative anchors based on GGIoU using the width and height of overlap regions between ground truths and predicting boxes to export the standard deviation of Gaussian distances. $\alpha$ = 0 means the baseline model using IoU to sampling.}
 \begin{tabular}{cc|cccccc} 
 \hline
$\beta$ & $\alpha$ & mAP & $\text{AP}_{50}$ &$\text{AP}_{60}$ & $\text{AP}_{70}$ &$\text{AP}_{80}$ &$\text{AP}_{90}$\\
 \hline
  - & 0   & 0.308 & 0.496 & 0.450 & 0.376 & 0.261 & 0.086\\
1/6 & 0.3 & 0.317 & 0.499 & 0.456 & 0.387 & 0.277 & 0.101\\
1/7 & 0.3 & 0.317 & 0.500 & 0.458 & 0.387 & 0.278 & 0.100\\
1/8 & 0.3 & 0.315 & 0.494 & 0.452 & 0.385 & 0.277 & 0.097\\
1/6 & 0.4 & 0.315 & 0.496 & 0.456 & 0.383 & 0.273 & 0.098\\
1/7 & 0.4 & 0.316 & 0.498 & 0.456 & 0.383 & 0.276 & 0.099\\
1/8 & 0.4 & 0.314 & 0.493 & 0.451 & 0.384 & 0.275 & 0.098\\
 \hline
 \end{tabular}
  \label{table:4.2}
\end{table}

\begin{table*}[t]
\begin{center}
\begin{tabular}{c|cc|cccccccccc}
\hline
method & $\gamma$ & ${\omega}_{loc}$ & mAP & $\text{AP}_{50}$ &$\text{AP}_{60}$ & $\text{AP}_{70}$ & $\text{AP}_{80}$ & $\text{AP}_{90}$ & $\text{AP}_{S}$ & $\text{AP}_{M}$ & $\text{AP}_{L}$\\
 \hline
baseline & - & 1.0 & 0.308 & 0.496 & 0.450 & 0.376 & 0.261 & 0.086 & 0.161 & 0.340 & 0.407\\
 \hline
IoU & 1.4 & 2.6 & 0.315 & 0.493 & 0.450 & 0.381 & 0.275 & 0.101 & 0.160 & 0.345 & 0.419\\
IoU & 1.3 & 2.5 & 0.316 & 0.496 & 0.452 & 0.384 & 0.276 & 0.104 & 0.168 & 0.345 & 0.423\\
IoU & 1.2 & 2.4 & 0.315 & 0.496 & 0.452 & 0.383 & 0.275 & 0.104 & 0.167 & 0.347 & 0.418\\
\hline
GGIoU & 1.5 & 2.9 & 0.317 & 0.494 & 0.451 & 0.386 & 0.279 & 0.107 & 0.166 & 0.345 & 0.426\\
GGIoU & 1.4 & 2.8 & 0.319 & 0.497 & 0.455 & 0.386 & 0.282 & 0.107 & 0.167 & 0.348 & 0.430\\
GGIoU & 1.3 & 2.6 & 0.319 & 0.500 & 0.457 & 0.388 & 0.278 & 0.104 & 0.165 & 0.348 & 0.432\\
GGIoU & 1.2 & 2.5 & 0.316 & 0.495 & 0.452 & 0.384 & 0.276 & 0.105 & 0.170 & 0.344 & 0.420\\
\hline
\end{tabular}
\\ \hspace*{\fill} \\
\caption{Comparison between IoU based and GGIoU based balanced localization function. $\gamma$ and ${\omega}_{loc}$ are the coupled hyper-parameters.}
\label{tab:4.3}
\end{center}
\end{table*}

\begin{table*}[t]
\begin{center}
\begin{tabular}{c|cc|cccccccccc}
\hline
method & $\gamma$ & ${\omega}_{loc}$ & mAP & $\text{AP}_{50}$ &$\text{AP}_{60}$ & $\text{AP}_{70}$ & $\text{AP}_{80}$ & $\text{AP}_{90}$ & $\text{AP}_{S}$ & $\text{AP}_{M}$ & $\text{AP}_{L}$\\
 \hline
baseline & - & 1.0 & 0.308 & 0.496 & 0.450 & 0.376 & 0.261 & 0.086 & 0.161 & 0.340 & 0.407\\
 \hline
IoU & 1.4 & 2.6 & 0.315 & 0.493 & 0.450 & 0.381 & 0.275 & 0.101 & 0.160 & 0.345 & 0.419\\
IoU & 1.3 & 2.5 & 0.316 & 0.496 & 0.452 & 0.384 & 0.276 & 0.104 & 0.168 & 0.345 & 0.423\\
IoU & 1.2 & 2.4 & 0.315 & 0.496 & 0.452 & 0.383 & 0.275 & 0.104 & 0.167 & 0.347 & 0.418\\
\hline
GGIoU & 1.5 & 2.9 & 0.317 & 0.494 & 0.451 & 0.386 & 0.279 & 0.107 & 0.166 & 0.345 & 0.426\\
GGIoU & 1.4 & 2.8 & 0.319 & 0.497 & 0.455 & 0.386 & 0.282 & 0.107 & 0.167 & 0.348 & 0.430\\
GGIoU & 1.3 & 2.6 & 0.319 & 0.500 & 0.457 & 0.388 & 0.278 & 0.104 & 0.165 & 0.348 & 0.432\\
GGIoU & 1.2 & 2.5 & 0.316 & 0.495 & 0.452 & 0.384 & 0.276 & 0.105 & 0.170 & 0.344 & 0.420\\
\hline
\end{tabular}
\\ \hspace*{\fill} \\
\caption{Comparison between IoU based and GGIoU based balanced localization function. $\gamma$ and ${\omega}_{loc}$ are the coupled hyper-parameters.}
\label{tab:4.3}
\end{center}
\end{table*}

\begin{table*}[t]
\begin{center}
\begin{tabular}{c|cccccccccc}
\hline
Backbone  & mAP & $\text{AP}_{50}$ &$\text{AP}_{60}$ & $\text{AP}_{70}$ & $\text{AP}_{80}$ & $\text{AP}_{90}$ & $\text{AP}_{S}$ & $\text{AP}_{M}$ & $\text{AP}_{L}$\\
 \hline
ResNet18 &0.308 & 0.496 & 0.450 & 0.376 & 0.261 & 0.086 & 0.161 & 0.340 & 0.407\\
ResNet50 & 0.356 & 0.555 & 0.510 & 0.432 & 0.311 & 0.113 & 0.200 & 0.396 & 0.468\\
ResNet101 & 0.377 & 0.575 & 0.533 & 0.460 & 0.337 & 0.130 & 0.211 & 0.422 & 0.495\\
ResNeXt101324d & 0.390 & 0.594 & 0.522 & 0.476 & 0.349 & 0.141 & 0.226 & 0.434 & 0.509\\
\hline
ResNet18* & 0.323 & 0.497 & 0.458 & 0.390 & 0.288 & 0.111 & 0.164 & 0.350 & 0.433\\
ResNet50* & 0.370 & 0.558 & 0.516 & 0.450 & 0.338 & 0.141 & 0.210 & 0.408 & 0.483\\
ResNet101* & 0.392 & 0.580 & 0.539 & 0.472 & 0.363 & 0.163 & 0.226 & 0.435 & 0.523\\
ResNeXt101324d* & 0.410 & 0.601 & 0.561 & 0.492 & 0.382 & 0.175 & 0.235 & 0.452 & 0.539\\
\hline
\end{tabular}
\\ \hspace*{\fill} \\
\caption{The results of balanced training method based on GGIoU. '*' means the method is used and the optimal parameters($\beta$ = 1/6, $\alpha$ = 0.3, $\gamma$ = 1.4) from ablation experiments are set. RetinaNet are trained with the proposal method with the image scale of [800, 1333] on COCO \textit{train-2017} and evaluated on \textit{val-2017}.}
\label{tab:4.4}
\end{center}
\end{table*}

\begin{table*}[t]
\begin{center}
\begin{tabular}{ccccccccc}
\hline
Model & Backbone & Schedule & mAP & $\text{AP}_{50}$ & $\text{AP}_{75}$ & $\text{AP}_{S}$ & $\text{AP}_{M}$ & $\text{AP}_{L}$\\
 \hline
YOLOv2~\cite{redmon2017yolo9000}& DarkNet-19 & - & 21.6 & 44.0 & 19.2 & 5.0 & 22.4 & 35.5\\
YOLOv3~\cite{redmon2018yolov3} & DarkNet-53 & - & 33.0 & 57.9 & 34.4 & 18.3 & 35.4 & 41.9\\
SSD300~\cite{Liu2016SSD} & VGG16 & - & 23.2 & 42.1 & 23.4 & 5.3 & 23.2 & 39.6 \\
SSD512~\cite{Liu2016SSD} & VGG16 & - & 26.8 & 46.5 & 27.8 & 9.0 & 28.9 & 41.9\\
Faster R-CNN~\cite{2017FasterRCNN} & ResNet-101-FPN & - & 36.2 & 59.1 & 39.0 & 18.2 & 39.0 & 48.2\\
Deformable R-FCN~\cite{deformable} & Inception-ResNet-v2 & - & 37.5 & 58.0 & 40.8 & 19.4 & 40.1 & 52.5\\
Mask R-CNN~\cite{he2017maskrcnn} & ResNet-101-FPN & - & 38.2 & 60.3 & 41.7 & 20.1 & 41.1 & 50.2\\
\hline
Faster R-CNN* & ResNet-50-FPN & 1x & 36.2 & 58.5 & 38.9 & 21.0 & 38.9 & 45.3 \\
Faster R-CNN* & ResNet-101-FPN & 1x & 38.8 & 60.9 & 42.1 & 22.6 & 42.4 & 48.5 \\
SSD300* & VGG16 & 120e & 25.7 & 44.2 & 26.4 & 7.0 & 27.1 & 41.5 \\
SSD512* & VGG16 & 120e & 29.6 & 49.5 & 31.2 & 11.7 & 33.0 & 44.2\\
RetinaNet* & ResNet-18-FPN & 1x & 31.5 & 50.5 & 33.2 & 16.6 & 33.6 & 40.1\\
RetinaNet* & ResNet-50-FPN & 1x & 35.9 & 55.8 & 38.4 & 19.9 & 38.8 & 45.0\\
RetinaNet* & ResNet-101-FPN & 1x & 38.1 & 58.5 & 40.8 & 21.2 & 41.5 & 48.2\\
RetinaNet* & ResNeXt-32x4d-101-FPN & 1x & 39.4 & 60.2 & 42.3 & 22.5 & 42.8 & 49.8\\
\hline
GGIoU SSD300 & VGG16 & 120e & 27.2 & 45.2 & 28.6 & 8.2 & 29.4 & 43.1\\
GGIoU SSD512 & VGG16 & 120e & 30.9 & 50.1 & 32.9 & 12.6 & 34.5 & 45.1 \\
GGIoU RetinaNet & ResNet-18-FPN & 1x & 32.8 & 50.6 & 35.3 & 17.7 & 34.6 & 41.9\\
GGIoU RetinaNet & ResNet-50-FPN & 1x & 37.3 & 56.1 & 40.1 & 20.9 & 40.1 & 46.4\\
GGIoU RetinaNet & ResNet-101-FPN & 1x & 39.5 & 58.5 & 42.7 & 22.1 & 42.7 & 50.2\\
GGIoU RetinaNet & ResNeXt-32x4d-101-FPN & 1x & 40.9 & 60.3 & 44.4 & 23.6 & 44.0 & 51.6\\
\hline
\end{tabular}
\\ \hspace*{\fill} \\
\caption{Comparisons of different models for accurate object detection on MS-COCO. All models are trained on COCO \textit{train-2017} and evaluated on COCO \textit{test-dev}. The symbol "*" means the reimplemented results in MMDetection\cite{chen2019mmdetection}. The training schedule "1x" and "120e" means the model is trained for 12 epochs and 120 epochs respectively.}
\label{tab:4.5}
\end{center}
\end{table*}

\begin{table*}[t]
\begin{center}
\begin{tabular}{cccccccc}
\hline
Model & Backbone & mAP & $\text{AP}_{50}$ & $\text{AP}_{60}$ & $\text{AP}_{70}$ & $\text{AP}_{80}$ & $\text{AP}_{90}$\\
 \hline
RetinaNet & ResNet-18-FPN & 47.5 & 74.9 & 69.7 & 58.3 & 39.9 & 13.9\\
RetinaNet & ResNet-50-FPN & 52.3 & 79.2 & 74.5 & 64.2 & 45.8 & 17.4\\
RetinaNet & ResNet-101-FPN & 54.5 & 79.7 & 75.8 & 66.2 & 50.7 & 21.0\\
RetinaNet & ResNeXt-32x4d-101-FPN & 56.0 & 80.8 & 77.1 & 67.4 & 52.3 & 23.0\\
\hline
GGIoU RetinaNet & ResNet-18-FPN & 49.4 & 74.4 & 70.5 & 60.2 & 43.9 & 17.3\\
GGIoU RetinaNet & ResNet-50-FPN & 54.2 & 78.7 & 74.9 & 65.9 & 49.8 & 21.6\\
GGIoU RetinaNet & ResNet-101-FPN & 56.4 & 79.9 & 76.4 & 68.3 & 53.0 & 24.6\\
GGIoU RetinaNet & ResNeXt-32x4d-101-FPN & 58.0 & 80.7 & 77.4 & 69.8 & 54.9 & 28.0\\
\hline
\end{tabular}
\\ \hspace*{\fill} \\
\caption{The results of RetinaNet are trained on the union of \textit{VOC2007 trainval} and \textit{VOC2012 trainval} and tested on \textit{VOC2007 test} with the image scale of [600, 1000] and different backbones.}
\label{tab:4.6}
\end{center}
\end{table*}

\begin{table*}[t]
\begin{center}
\begin{tabular}{cccccccccccc}
\hline
Backbone & mAP & $\text{AP}_{50}$ & $\text{AP}_{55}$ & $\text{AP}_{60}$ & $\text{AP}_{65}$ & $\text{AP}_{70}$ & $\text{AP}_{75}$ & $\text{AP}_{80}$ & $\text{AP}_{85}$ & $\text{AP}_{90}$ & $\text{AP}_{95}$\\
 \hline
ResNet18 & 78.0 & 94.6 & 93.8 & 92.8 & 91.5 & 89.3 & 86.4  & 82.3 & 75.4 & 59.6 & 14.0\\
ResNet50 & 80.7 & 95.4 & 95.0 & 94.2 & 93.0 & 91.3 & 88.7 & 84.9 & 78.6 & 65.2 & 21.0\\
ResNet101 & 82.7 & 96.0 & 95.4 & 94.9 & 93.8 & 92.2 & 90.1 & 87.1 & 81.8 & 70.0 & 25.9\\
ResNeXt-32x4d-101 & 83.2 & 96.2 & 95.8 & 95.1 & 94.2 & 92.7 & 90.4 & 87.3 & 82.8 & 70.7 & 26.4\\
\hline
ResNet18* & 81.2 & 95.2 & 94.8 & 93.9 & 92.8 & 91.0 & 88.6 & 85.0 & 79.7 & 67.1 & 23.7\\
ResNet50* & 84.0 & 96.2 & 95.6 & 95.0 & 94.2 & 92.6 & 90.6 & 87.6 & 83.2 & 72.8 & 32.6\\
ResNet101* & 85.6 & 96.6 & 95.8 & 95.6 & 94.7 & 93.6 & 91.8 & 89.4 & 85.1 & 76.2 & 37.3\\
ResNeXt-32x4d-101* & 86.1 & 96.8 & 96.5 & 95.9 & 95.0 & 94.0 & 92.0 & 89.9 & 85.7 & 76.4 & 38.4\\
\hline
\end{tabular}
\\ \hspace*{\fill} \\
\caption{The main results of RetinaNet is trained on \textit{SCD-train} and tested on \textit{SCD-test} with the image scale of [800, 1333] and different backbones.}
\label{tab:4.7}
\end{center}
\end{table*}

\subsection{Experiment Setting}
\textbf{Dataset.}
All detection baselines are trained and evaluated on three object detection datasets including MS COCO~\cite{lin2014MSCOCO}, PASCAL VOC~\cite{Everingham2010PascalVOC} and SCD\cite{yang2021scd}. MS COCO contains about 118K images for training(\textit{train-2017}), 5k images for validation(\textit{val-2017}) and 20k for testing(\textit{test-dev}). For PASCAL VOC, the VOC2007 includes 5011 training images(\textit{VOC2007 trainval}) and 4952 testing images(\textit{VOC2007 test}) while VOC2012 contains 11540 training images and 10591 testing images. The combination of \textit{VOC2007 trainval} and \textit{VOC2012 trainval} are adopted to train models while VOC2007
(\textit{VOC2007 test}) is used to evaluate models. SCD is a dense carton detection and segmentation dataset and contains LSCD and OSCD subsets. We conduct experiments on LSCD which contains 6735 image for training and 1000 images for testing.

\textbf{Evaluation Protocol.} We adopt the same evaluation metric as MS COCO 2017 Challenge~\cite{lin2014MSCOCO}. This includes AP(averaged mAP across different IoU thresholds $IoU = \{.5, .55, \cdots, .95\}$), $\text{A}{{\text{P}}_{50}}$(AP at IoU threshold of 0.5), $\text{A}{{\text{P}}_{75}}$(AP at IoU threshold of 0.75), $\text{A}{{\text{P}}_{S}}$ (AP for small scales), $\text{A}{{\text{P}}_{M}}$ (AP for medium scales) and $\text{A}{{\text{P}}_{L}}$ (AP for large scales).

\textbf{Implementation Details.} All experiments are implemented by using PyTorch and MMDetection \cite{chen2019mmdetection}. All models are trained on 4 NVIDIA GeForce 2080Ti GPU while the learning rate changes linearly based on the mini-batch \cite{goyal2017accurate}. For all ablation studies, RetinaNet with ResNet18 as backbone are trained on \textit{train-2017} and evaluated on \textit{val-2017} with image scale of [800, 1333]. For all baseline models, we follow the default settings of MMDetection.

\subsection{Ablation studies on using GGIoU as sampling metric}
\label{4.2}

\textbf{Ablating the style of calculating the standard deviation of Gaussian distance.} As shown in Table~\ref{table:4.1}, when using the width and height of truth boxes to export the standard deviation $\sigma_{1}$ and $\sigma_{2}$ of Gaussian distances, the sampling strategy based on GGIoU can improve the model's AP by up to 0.5\%. Table~\ref{table:4.2} reports the results that when the overlapping area between ground truth boxes and prediction boxes acts as base size to derive standard deviations, 0.9\% mAP is boost higher 0.4\% than the base size of IoU counterpart. These results echo the argument in Section~\ref{section 3.1} and verify that employing the shape of the overlapping area between the truth boxes and the prediction boxes can not only measures the distance of two boxes but reflects the overlapping scale which makes reasonable.

\textbf{The analyse of improvements for localization accuracy based on using GGIoU as sampling metric.} Table~\ref{table:4.2} highlights that the performance of detector is improved from 30.8\% to 31.7\%. With $0.3\% \sim 0.6\%$ improvements in ${AP}_{50} \sim {AP}_{60}$ and $1.1\% \sim 1.6\%$ increases in ${AP}_{70} \sim {AP}_{90}$, it reveals that the sampling strategy based on GGIoU is benefit to facilitate localization. It attributes to paying more attention to the features which receptive field is better aligned with the objects. This favor encourage model to touch the precise boundary. 

\begin{figure}[!tb]
\begin{center}
\includegraphics[width=\linewidth]{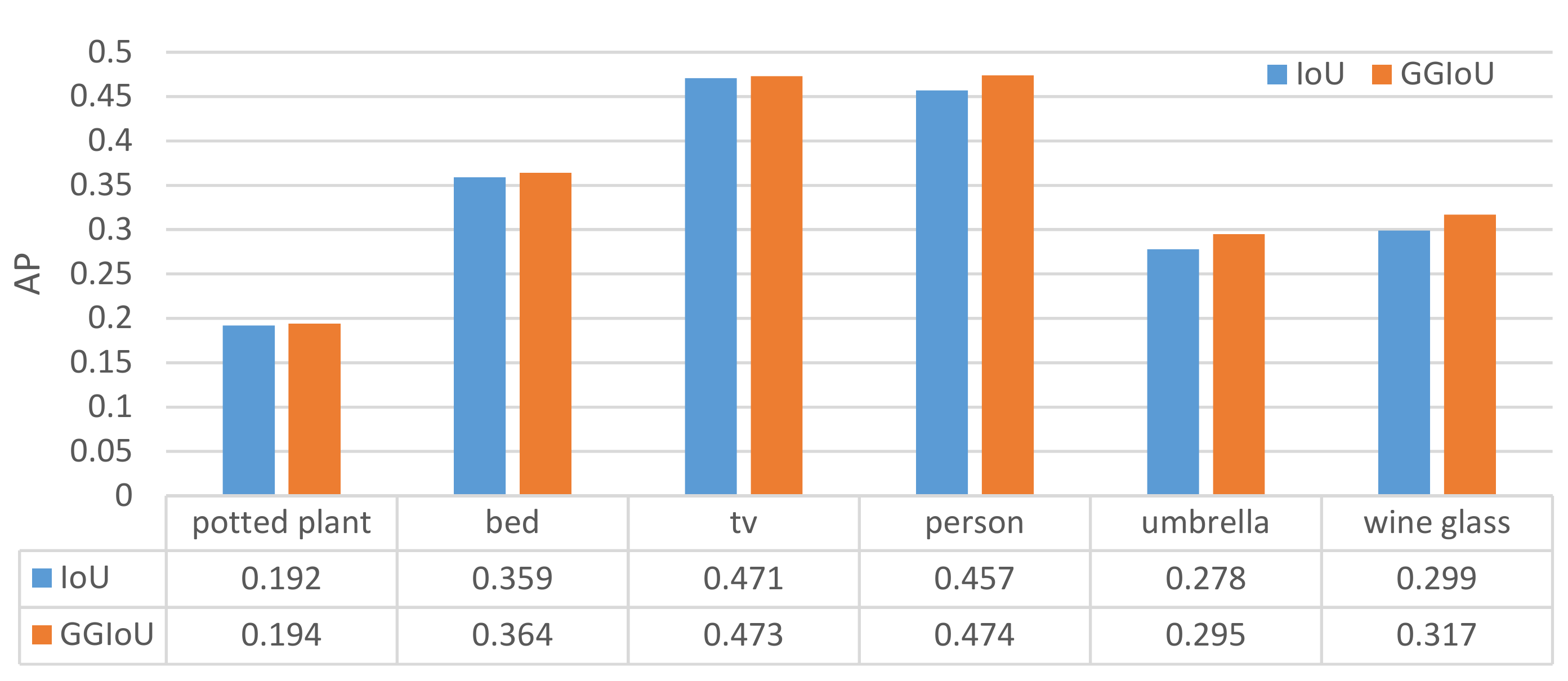}
\caption{Comparison between IoU based and GGIoU based sampling strategy with respective to slender objects.}
\label{category}
\end{center}
\end{figure}

\begin{figure}[!tb]
\begin{center}
\includegraphics[width=\linewidth]{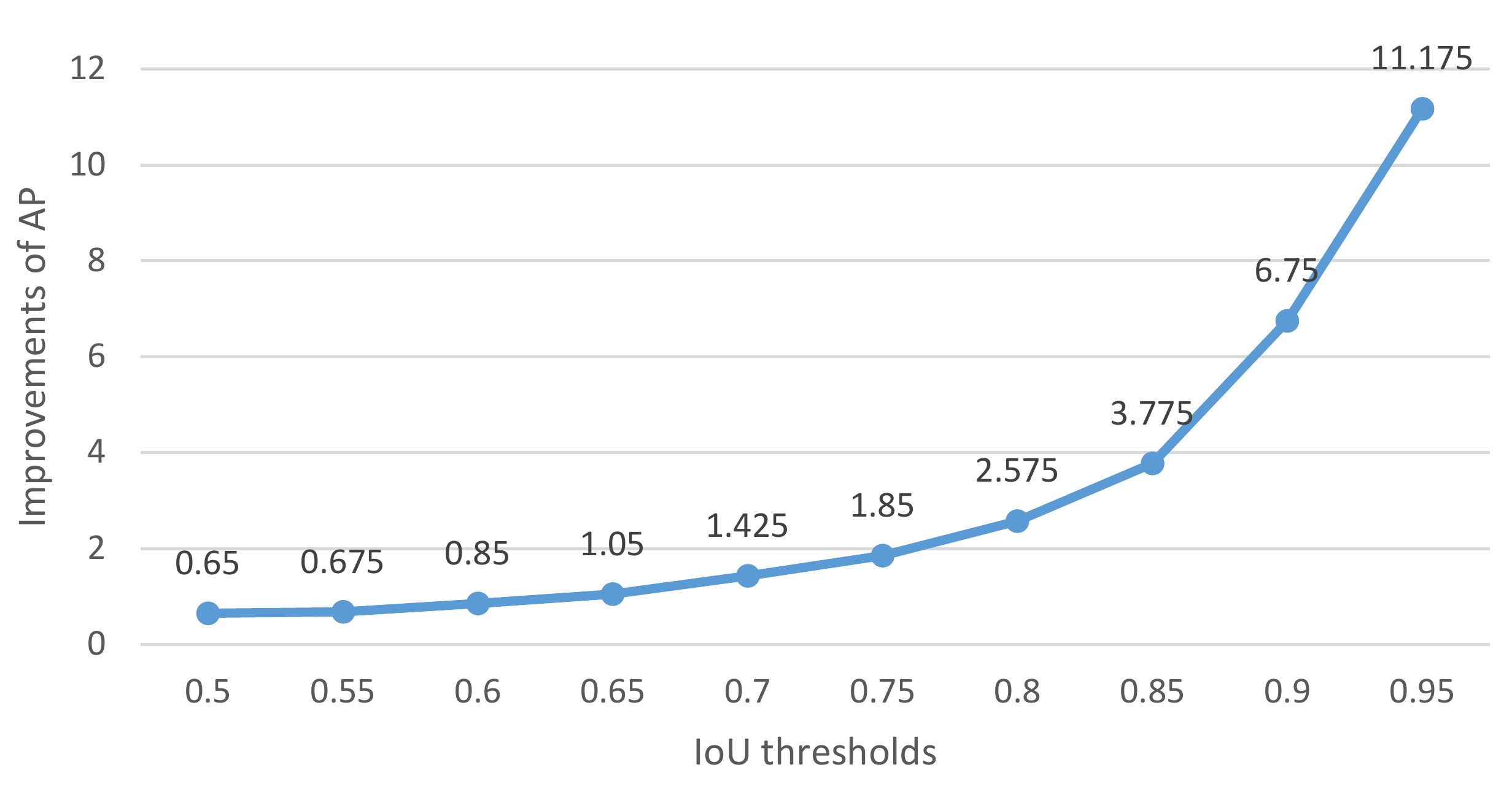}
\caption{The average improvements of ResNet18, ResNet50, ResNet101 and ResNeXt-32x4d-101 with the increases of AP's thresholds on SCD.}
\label{SCD imprivements}
\end{center}
\end{figure}

\textbf{The analyse of improvements for slender objects based on using GGIoU as sampling metric.} As shown in Figure~\ref{category}, we list the performance improvements with respect to some non-slender objects, such as potted plant, bed and tv, decent gains are achieved by $0.2\% \sim 0.5\%$ when using our proposal GGIoU to assign positive and negative samples. For slender objects such as person, umbrella and wineglass, significant improvements are achieved by $1.7\% \sim 1.8\%$. The results verify the thesis in Section~\ref{section 3.2} and reveal that slender objects harvest more positive samples instead of only one, balancing the attention to all objects and enhance the overall performance.

\subsection{Ablation experiments for localization loss function based on GGIoU as weighting scare.} For simplicity, the parameters of GGIoU directly adopt the optimal values by ablating in Table~\ref{table:4.2}, setting $\sigma$ = 1/6 and $\alpha$ = 0.3. In Eq.\ref{eq:GGIoU-balancedLoc} and Eq.\ref{eq:GGIoU-balancedLocPara}, IoU is replaced by GGIoU to calculate weights for positive samples.

\textbf{Ablating the controlling parameters.} As shown in Table~\ref{tab:4.3}, when $\gamma$ is set as 1.4, the proposal localization loss function balanced by GGIoU achieves best localization accuracy. Without adding bells and whistles, it improves the AP of baseline by 1.1\% on MS COCO \textit{val2017} while powerful gains are achieved by $1.0\% \sim 2.1\%$. In addition, decent gains are achieved in ${AP}_{50}$ and ${AP}_{60}$ by 0.1\% and 0.5\% respectively. The experiments indicate that the GGIoU-balanced localization function can effectively promote localization performance.

\textbf{Comparison between GGIoU and IoU on balanced localization function.} As shown in Table~\ref{tab:4.3}, using IoU to balance localization function with best controlling parameter can improve the AP by 0.8\%. Through comparison, adopting GGIoU can further boost performance by 0.3\% which verifies the analyse in Section~\ref{section 3.3}. The improved metric not only inherits the merit of balanced localization function which relieves the learning gradient is dominated by abnormal and noisy outliers, but pays more attention to features equipped with receptive field aligned better with objects, which is conducive to coordinate regression.

\subsection{Experiments for GGIoU-balanced learning Method.} We combine the sampling strategy and balanced localization function based on GGIoU and call this union method as GGIoU-balanced training method. Table~\ref{tab:4.4} reports that the union method yields considerable performance by $1.4\% \sim 2.0\%$ on RetinaNet with different backbones which reveals that the two proposal method based GGIoU can complement each other and shows the superiority of GGIoU. Meanwhile, the combined method consistently boosts the performance of ${AP}_{50} \sim {AP}_{60}$ by $0.1\% \sim 0.9\%$ AP and ${AP}_{70} \sim {AP}_{90}$ by $1.2\% \sim 3.4\%$ AP which affirms it is versatile. Finally, all scare of backbones embedded into RetinaNet with our method achieve decent gains of performance that shows great robustness. Albeit notable gains by ResNeXt, these is still a powerful boost by 2.0\%.

\subsection{Main Results}

To reveal the versatility and generalization of GGIoU-balanced learning strategy on different scenarios, we conduct experiments on general object detection dataset such as MS COCO and PASCAL VOC, as well as a large-scale stacked carton dataset such as SCD.

\textbf{Main results on MS COCO.} Table~\ref{tab:4.5} reports the results of GGIoU-balanced learning strategy on COCO \textit{test-dev} compared with some advanced detectors. The proposal method embedded into RetinaNet can improve the AP by $1.3\% \sim 1.5\%$ while significant gains are achieved in ${AP}_{75}$ by $1.7\% \sim 2.1\%$. As for SSD counterpart, $1.3\% \sim 1.5\%$ improvements are also gained in AP and $1.7\% \sim 2.2\%$ increases are yielded in ${AP}_{75}$. Therefore, it is concluded that anchor-based single shot detector equipped with the strategy method can yield decent improvements in challenging general object detection scenarios.

\textbf{Main results on PASCAL VOC.} To verify the generalization on general object detection dataset, we train RetinaNet with our method on PASCAL VOC and compare with the baseline. As shown in Table~\ref{tab:4.6}, our proposal method brings $1.9\% \sim 2.0\%$ improvements in AP and $2.3\% \sim 5.0\%$ increases in ${AP}_{80}$/${AP}_{90}$. The gain effects are the same as MS COCO, which indicates that the proposal approach can work well in different general object detection dataset scenarios.

\textbf{Main results on SCD.} We conduct experiments on SCD so as to show the performance on industrial application scenarios. RetinaNet fed the image scale of [800, 1333] with different backbones are trained on \textit{LSCD-train} and tested on \textit{LSCD-test} with the same image scale of [800, 1333]. As shown in Table~\ref{tab:4.7}, significant gains are achieved by $2.9\% \sim 3.3\%$ in AP. The localization accuracy boost are detailed in Figure~\ref{SCD imprivements}, the accuracy is continuously boost with the increase of the AP threshold from 0.65\% to 11.175\%. It indicates that the method can be applied to industrial scenarios without costing inference overhead.
\section{Conclusion}
IoU lacks the ability to perceive the center distance between ground truths and the bounding box of objects, which damages the performance of anchor-based detectors during training. In this paper, we introduce a new metric, namely GGIoU, which can pay more attention to the center distance among bounding boxes. Based on GGIoU, a balanced training method is proposed, consisting of GGIoU-based sampling strategy and GGIoU-balanced localization function. The union method can not only ease the dilemma of slender objects which are only assigned one positive sample, but guide model pay more attention to the feature whose receptive field better align with the region of objects. The property help improve localization accuracy and general performance. The proposed method shows powerful effect and generalization  on general object detection dataset such as MS COCO and PASCAL VOC as well as large scale industrial dataset such as SCD.


{\small
\bibliographystyle{ieee}
\bibliography{ref}
}

\end{document}